# Weakly Supervised Arrhythmia Detection Based on Deep Convolutional Neural Network

Yang Liu[1], Kuanquan Wang[1], Qince Li[1,2], Runnan He[1,2], Yongfeng Yuan[1], and Henggui Zhang[2,3,4]

[1] School of Computer Science and Technology, Harbin Institute of Technology, Harbin, China
[2] Peng Cheng Laboratory, Shenzhen, China
[3] School of Physics and Astronomy, University of Manchester, Manchester, UK
[4] International Laboratory for Smart Systems and Key Laboratory of Intelligent of Computing in Medical Image, Ministry of Education, Northeastern University, Shenyang, China

H.Zhang-3@manchester.ac.uk

*Abstract*—Supervised deep learning has been widely used in the studies of automatic ECG classification, which largely benefits from sufficient annotation of large datasets. However, most of the existing large ECG datasets are roughly annotated, so the classification model trained on them can only detect the existence of abnormalities in a whole recording, but cannot determine their exact occurrence time. In addition, it may take huge time and economic cost to construct a fine-annotated ECG dataset. Therefore, this study proposes weakly supervised deep learning models for detecting abnormal ECG events and their occurrence time. The available supervision information for the models is limited to the event types in an ECG record, excluding the specific occurring time of each event. By leverage of feature locality of deep convolution neural network, the models first make predictions based on the local features, and then aggregate the local predictions to infer the existence of each event during the whole record. Through training, the local predictions are expected to reflect the specific occurring time of each event. To test their potentials, we apply the models for detecting cardiac rhythmic and morphological arrhythmias by using the AFDB and MITDB datasets, respectively. The results show that the models achieve beat-level accuracies of 99.09% in detecting atrial fibrillation, and 99.13% in detecting morphological arrhythmias, which are comparable to that of fully supervised learning models, demonstrating their effectiveness. The local prediction maps revealed by this method are also helpful to analyze and diagnose the decision logic of record-level classification models.

## I. Introduction

Cardiac arrhythmias, manifested as abnormal heart rates or disturbed rhythms, have become the leading causes of morbidity and mortality worldwide [1]. Electrocardiogram (ECG) is still the most common diagnostic tool for arrhythmias due to its safety, reliability and economics. With the development of communication technology and miniaturization of ECG devices, wearable ECG monitoring services are becoming more popular practical uses. However, the manual analysis of ECG recordings is time-consuming and error-prone, constituting a challenge to process big volume of data with an acceptable time and economic cost. Therefore, automatic ECG analysis technology is in demand to cope with this challenge.

In recent years, the deep learning technology (aka, deep neural network, DNN) has greatly promoted the development of automatic ECG analysis, especially automatic ECG classification. Some studies claim to have model performance better than that of average cardiologists on detection of a wide spectrum of ECG abnormalities [2]. The success of DNN in ECG classification is largely benefited from the large-scale annotated ECG datasets, such as PhysioNet/CinC challenge 2017 dataset [3], China physiological signal challenge 2018 dataset [4]. Generally, these datasets just contain annotations about the rhythm types existing in a whole record, but no specific time information of their occurrence of each arrhythmic episode. As a result, the models trained with these datasets are designed just to identify the types of rhythm occurred in a record, but not to specify their occurring time. However, the occurrence time of ECG abnormalities is very important to the diagnosis, especially when the recording is quite long (over 48 hour's recording for Holter monitor). Even for a short recording, the detected occurrence time of each rhythm can also help to uncover the decision logic of the DNN model, and assist in model diagnosis and improvement. However, fully supervised learning method for this problem requires fine labeling of the occurrence time of each cardiac event, which would entail huge labor and time costs. Besides, the large amount of electronic medical record data accumulated by medical institutions are basically roughly annotated. How to make full use of these data resources is still an open question.

In this study, we propose a weakly supervised learning (WSL) method for arrhythmias detection in order to reduce the dependency of the model on training by fine labeled dataset. During the model training, the new method just uses available labels of an ECG recording to get information about the absence/presence of heart rhythms in the recording. But, by constraining the decision-making process of the model, the model is forced to infer for itself the exact timing of each rhythm. The decision process of the DNN model is divided structurally into two stages, namely local prediction and global aggregation. In the local prediction stage, the model's output (hidden in the network) is a sequence of vectors, each indicating probabilities of rhythms in a local time period. And then, in the global aggregation stage, the local predictions are aggregated into a single vector in which each element indicates the probability for a corresponding rhythm to exist in the whole record. Therefore, we can get the occurrence time information of rhythms from the local predictions of the model, although the model training isn't supervised by the time information.

The main contributions of this work are summarized as follows:

1) We propose weakly supervised detectors for cardiac arrhythmia that can detect the exact occurrence time of arrhythmias but are trained using only coarse record-level labels. To our knowledge, this is the first publicly accessible work that deals with the problem of arrhythmia detection in a weakly supervised manner.
2) We design special aggregation mechanisms for rhythmic arrhythmias and morphological arrhythmias, and achieve performances comparable to the state-of-the-art of fully supervised methods on the AFDB and MITDB databases.
3) Our proposed models accept data with variable input lengths, and are more efficient than models that make predictions by using sliding windows.
4) The local prediction maps generated by the proposed models can be used to reveal the patterns recognized by the model, which is helpful to interpret and diagnose the DNN model even though the purpose is only record-level classification.

The remainder of this paper is organized as follows. In section II, we introduce related works, including weakly supervised learning and cardiac arrhythmia detection. Section II describes our WSL method for arrhythmia detection in details. Section III presents a series of experimental results to evaluate the performance of our method. In section IV, we discuss the effectiveness and limitation of the proposed method, and discuss potential future research issues. Finally, we conclude this work in section V.

## II. Related Works

Computer-aid diagnosis of cardiac arrhythmia has been studied for several decades, and in recent years, it has been remarkably boosted by the deep learning technology. In this work, we also follow the methodology of deep learning to deal with this problem, but in a weakly supervised approach. Although the concept of weak supervised learning has not been applied in the field of ECG analysis, it has been studied in the field of computer vision (CV), especially for purposes of object localization and semantic segmentation. We will introduce some relevant works regarding arrhythmia detection and weakly supervised learning in this section.

### A. Cardiac Arrhythmia Detection

From the manifestations in ECG, cardiac arrhythmias can be categorized as morphological arrhythmia or rhythmic arrhythmia. Morphological arrhythmia is manifested as an abnormal heart beat in ECG, e.g., premature ventricular contraction (PVC), while rhythmic arrhythmia is formed a series of heart beats with irregular intervals, e.g., atrial fibrillation (AF). Research works based on deep learning has been published for detection of both types of arrhythmias, and some have achieved significant improvements in performance compared with the traditional methods [5-7].

For detection of morphological arrhythmias, the input of a DNN model is typically a short ECG segment containing just the heartbeat that need to be classified or also the heartbeats that are adjacent to it. A diversity of DNN architectures have been developed for this task. Among them, convolutional neural network (CNN) and its variants are most frequently used in this domain [8-10]. CNN can model the structure of data in its receptive field with relatively few parameters [11]. Besides, CNN is easy to be processed in parallel, which makes it have good acceleration effect on GPU and other massively parallel processors. The DNN architectures that have been applied on this task also include restricted Boltzmann machine (RBM) [12] and recurrent neural network (RNN) [13]. The above methods almost all rely on other QRS detection methods to locate the heartbeats, but a novel U-net based DNN proposed by [14] solved the problems of QRS detection and heartbeat classification simultaneously.

When the goal is to detect rhythmic arrhythmias, the input of the model is usually not the ECG signal for a single heartbeat, but a long segment with a length ranging from a few seconds to tens of seconds as the rhythmic arrhythmias generally span a long time period. Some architectures that are developed for morphological arrhythmia detection can also be used for detection of rhythmic arrhythmias. CNN is also the most commonly used architecture for this task, but usually needs a wider receptive field [6, 15]. Therefore, many researchers have introduced new mechanisms to make the CNN have deeper structure and stronger recognition capability. A deep 1D residual neural network (1D ResNet) with up to 34 convolutional layers was first introduced by [2] for detection of 12 types of cardiac rhythms and achieved high diagnostic performance similar to that of cardiologists. 1D ResNet has also been combined with RNN [16] and attention mechanisms [17] for further performance improvements.

### B. Weakly Supervised Learning

The technology of WSL has been intensively studied for object localization and semantic segmentation in the field of computer vision, where the training process is just supervised by the existence of object instances in the image. The solution of this problem will obviate the need for large-scale fine-annotated datasets, thus can save lots of labor and time costs for labeling. However, in the field of ECG analysis, although a large number of clinical test records can be potentially leveraged for WSL just with little effort of annotation, to the best of our knowledge, the WSL method hasn't been studied for detection of ECG abnormalities.

CNNs have been widely studied for weakly supervised learning [18-20]. The architecture of CNN is basically constituted by a stack of pattern searchers, and its intermediate outputs represent the spatial distribution of features of a certain level, i.e., feature maps. Therefore, CNN has the potential to locate the objects despite be trained on image-level labels. In order for the network to be trained end-to-end, a mechanism is necessary to link the local predictions, i.e., the feature maps, to the image-level labels. The mechanism will determine the distribution of gradients on the feature map during the model training, and thus it has a very important impact on the localization performance. In many studies, the mechanism is implemented by aggregating the local predictions into the global prediction, e.g., global average pooling (GAP) [21] and global maximum pooling (GMP) [18, 22]. Some researchers also proposed mediate mechanisms between GAP and GMP, such as Log-Sum-Exp (LSE) [19] and global weighted rank pooling (GWRP) [23]. Besides the aggregation mechanism, the image-level supervision can also be transferred to constraints on the feature maps. For example, researchers in [24] proposed constrained CNN (CCNN) in which a generalized constraint mechanism is capable to model any linear constraints on the predicted label distribution, such as presence, absence and proportion of each label.

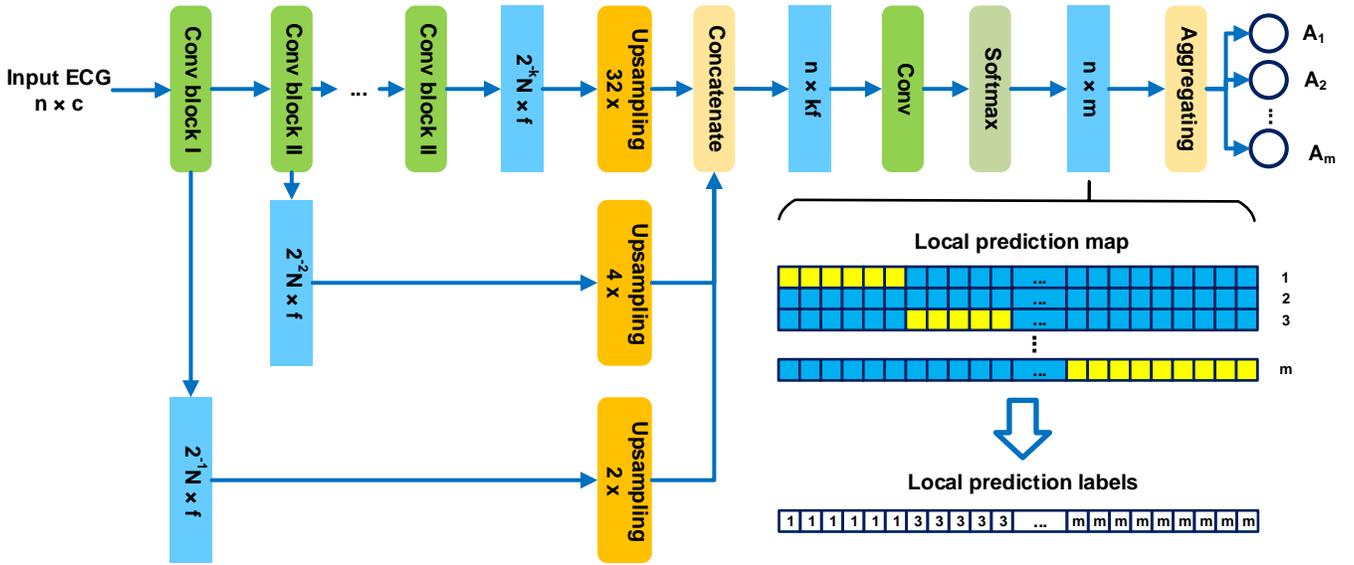

Fig. 1. Overview of our proposed WSL framework for arrhythmia detection. Given an input ECG is in a form of (n, c) with n denoting the number of the sampling points and c the number of the recording channels, we forward propagate the ECG through a stack of convolutional blocks for feature extraction. A feature map that includes features of different scales is obtained by upsampling and concatenating the outputs of each convolutional blocks. Then, based on the feature map, one or several convolutional layers with kernel size of 1 and a softmax activation layer is used to generate the local prediction map, where each column is the predicted probability distribution of arrhythmias on its corresponding position. Finally, the local prediction map is aggregated to produce the whole prediction for the record, which is a vector representing the probability of each rhythm appearing in the signal. With the supervised information of rhythm existence, we compute the loss of whole predictions and optimize the network by back propagation of loss gradients. In this figure, boxes with round corners represent the process, while boxes with square corners represent the intermediate data.

### III. Methods

*A. Problem Formulation*

We define an ECG recording as a series of sampling points $X = \{x_1, ..., x_n\}$ where *n* is the number of sampling points in the record. The goal is to construct a model *D* that generates labels $Y = \{y_1, ..., y_n\}$ for an input *X*. $y_i \in L = \{1, ..., m\}$ denotes the rhythm type of $x_i$, and *L* is the set of considered rhythm types. Here, we consider *D* as a parametric probabilistic model which generates a probability distribution of $y_i$ for each $x_i$: $D(X; \theta)_{i,j} = p(y_i=j \mid X)$, where $\theta$ is the parameters of *D*. To find a suitable $\theta$, we need to train the model on a dataset. Ideally, the training set provides the ground truth of *Y* for each *X* in it. However, in this study, we try to solve the problem in a weakly supervised setting: each training example is a tuple $(X, L_X)$ in which $L_X \subset L$ only indicates the rhythm types present in *X*.

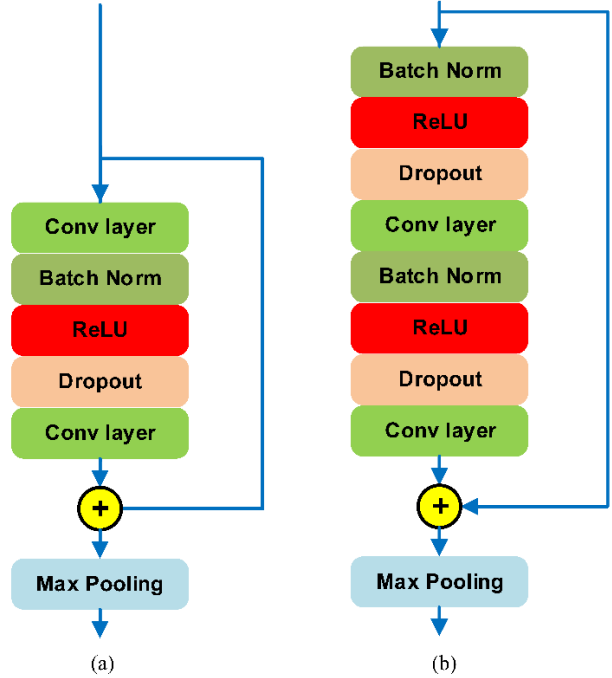

Fig. 2. Structure of convolutional blocks. (a) Type I convolutional block. (b) Type II convolutional block.

The above formulation is applicable to both tasks of detecting rhythmic and morphological arrhythmias. However, when considering that detection of morphological arrhythmias is at the heartbeat level, we can extend the above formulation to the case in which heartbeat is the location time unit for the detection. In this situation, the target labels for $X$ is $Y_B = \{y_1, \ldots, y_b\}$ where $b$ is the number of heartbeats in $X$, and $y_i$ is the rhythm type of the $i^{th}$ heartbeat. As the heartbeat in an ECG recording is orders of magnitude less than the sampling points, the space of the possible solutions for $Y_B$ is far smaller than that for $Y$. Therefore, this transformation significantly simplifies the original problem.

*B. Base Architecture*

We propose a novel DNN architecture for weakly supervised arrhythmia detection, as shown in Fig. 1. The main body of this architecture is a 1D ResNet, which has been demonstrated effective in the supervised arrhythmias detection [16, 17]. To make it suitable for weakly supervised learning, we make two important extensions to it: one is local prediction before aggregation, the other is concatenation of features of different scale. We will explain both of the extensions in the following.

The input of the model is an ECG signal in a form of $n \times c$, where $n$ is the number of sampling points and $c$ is the number of channels (or leads) for the signal. The input is processed by a stack of convolutional blocks, each belongs to one of two types: type I and type II. The structures of both types of the convolutional blocks are shown in Fig. 2. Each type of block contains two convolutional layers and other assistant layers. The output of a block's last convolutional layer is merged with its input by element-wise addition, then a max pooling layer with a pool size of 2 compresses the output to half of its original length. The difference of the two types is that the input of a type II block is processed by 3 layers (i.e., batch normalization, ReLU activation and dropout) before entering to its first convolutional layer, while in a type I block, these 3 layers are omitted. In the whole architecture, only the first convolutional block is in type I, the others are all of type II. Due to the max pooling layers, the feature map shrinks as processed by the blocks. For example, feature map outputted by the first block is in a form of $n/2 \times f$, where $n/2$ is the length of the feature map, and $f$ is the number of channels of the feature map (different from the number of channels of the model input, i.e., $c$) which is determined by the number of filters in the convolutional layer. As a result, the resolution of the final block output is much

6lower than that of the input, and thus affect the precision of local predictions. To solve this problem, we upsample the output of each block to the length of the input by simply repeating its elements, then concatenate the outputs from different blocks along the channel dimension. The mechanism is similar to the layer fusion in [25], but the difference is that, in this mechanism, features at different levels are concatenated rather than added up. This design makes features of different levels accessible to the local prediction layer, and enable the local prediction layer to learn a logic that selects features at the appropriate levels for the prediction.

After the concatenation, we get a feature map in a form of $n \times kf$, where $k$ is the number of convolutional blocks, and $kf$ is the number of channels (or features) of the feature map. Then, we make local predictions base on it by using convolutional layer and softmax activation. The number of filters in the convolutional layer is $m$, i.e., the number of classes (or label types) considered by the model. The kernel size of each filter is only 1, which means that the prediction is made with only a single feature vector of the feature map. Then softmax activation is applied along the channel axis, which generates the probability distribution of labels for each sampling point in the input. All the probabilities formed a local prediction map in a shape of $n \times m$, as shown in Fig. 1. Given a local prediction map, we can easily get the local prediction labels, i.e., $Y$ for $X$ in the formulation, by finding the position of maximum activation value at each sampling point.

To make the model trainable with record-level labels, we add an aggregation layer after the local predictions to generate record-level predictions. This is one of the main differences of our model with the general supervised ones which typically first aggregate local features to global features and then make record-level predictions based on the global feature vector. Here, we restrict the aggregation to be done along the time axis, and the output is a vector with length of $m$ indicating the predicted probability of the existence of each rhythm. The aggregation mechanisms are described in detail in the next subsection. The loss of the prediction can be computed by the loss functions that are designed for supervised record-level prediction models. We note that the loss function should support multi-label classifications, as multiple rhythms may coexist in a single ECG record. Therefore, we use the binary cross entropy as the loss function for training of our models.

*C. Aggregation Mechanisms*

The aggregation mechanism is critical to the success of a DNN model for weakly supervised learning. In the field of computer vision, different methods have been proposed, such as GAP, GMP and LSE. GAP averages all the local predictions as in (1):

$$S_c = \frac{1}{n}\sum_{i=1}^{n} D_{i,c}, \tag{1}$$

where $D$ is a shorthand of $D(X; \theta)$. We argue that GAP is not suitable for use in the weakly supervised arrhythmia learning, as some arrhythmias just self-terminate in a very short time period of an ECG record. For example, premature ventricular contractions (PVC) are generally very rare in a ECG record, and are mostly isolated. When an ECG record contains only one PVC, the application of GAP will cause the activation value for the PVC too low and mislead the gradient during training.

In contrast, GMP selects only the maximum value for a class as its record-level activation as in (2):

$$S_c = \max_{i \in \{1,\dots,n\}} D_{i,c}. \tag{2}$$

By using GMP, a rhythm is considered to be present in a record as long as it occurs at some time in the record regardless of duration. One possible problem of GMP is that it only back propagates the gradient to the position that has the maximum activation. When an arrhythmia exists in a record, it seems that the gradient for the loss of its detection will vanish as long as one of the activation values for the arrhythmia is close to 1 during the whole record. For this reason, some researchers suggests that GMP can lead to underestimation of the true region of an object [23]. However, in our architecture, softmax makes the activations of all classes sum to 1 at any position. Consequently, the loss from classes that are not present in a record will force the activation values to be transferred to the classes that do exist in the record. Therefore, we argue that GMP doesn't cause underestimation of the period of a rhythm, and thus may be a practical aggregation mechanism for the proposed architecture.

The aggregation can also be done in a mechanism between GAP and GMP. For example, LSE is a convex approximation of the max function [19]:

$$S_c = \frac{1}{r}\log(\frac{1}{n}\sum_{i=1}^{n}\exp(rD_{i,c})) ,  \quad (3)$$

where $r > 0$ is a hyper-parameter that controls the degree of the approximation to GMP. When $r$ is in a high value, this function is similar to GMP, but when $r$ is very low, this function is similar to GAP.

All the mechanisms described above, which we call ordinary aggregation mechanisms in this paper, just focus on the activation values, and are not discriminatory for positions. In fact, it is unnecessary to make detection at every sampling point of an ECG recording. As mentioned above, morphological arrhythmias are generally detected in unit of heartbeat, thus we can select only one prediction in each heartbeat for the aggregation. Specifically, we mask out all the local predictions except the prediction at the R peaks which are in the middle region of the corresponding heartbeats and can usually be accurately recognized by algorithms such as the Pan-Tompkins algorithm [26]. On the selected local predictions, the aggregation methods mentioned above can also be applied. Then, we get the corresponding masked aggregation mechanisms: global average R-peak predictions pooling (GMAP) as in Equation (4), global maximum R-peak predictions pooling (GMRP) as in Equation (5), and Log-Sum-Exp on R-peak predictions (LSER) as in Equation (6).

$$S_c = \frac{1}{|B|}\sum_{i \in B} D_{i,c} ,  \quad (4)$$

$$S_c = \max_{i \in B} D_{i,c} ,  \quad (5)$$

$$S_c = \frac{1}{r}\log(\frac{1}{|B|}\sum_{i \in B}\exp(rD_{i,c})) ,  \quad (6)$$

where $B$ denotes the set of R peak positions. These mechanisms lead the model to learn features around the R peak, and thus can reduce the difficulty of pattern recognition. Furthermore, the number of local predictions that are used for aggregation are significantly reduced, which can help to simplify the problem of weakly supervised arrhythmia detection for reasons discussed in the section of problem formulation.

## IV. Experiments

We evaluate the proposed methods on the detection tasks of rhythmic and morphological arrhythmias. In addition to the evaluation of the overall architecture, we also assess the effects of different aggregation methods, different model depth and different data distribution on the performance of the models.

### A. Datasets

Atrial fibrillation (AF), as one of the most common rhythmic arrhythmias, is chosen in our experiment as the object to be detected in the task of rhythmic arrhythmias detection. The model for this task is evaluated on the MIT-BIH Atrial Fibrillation Database (AFDB) [27, 28]. For convenience of model training and testing, we split the recordings into segments of 20 seconds. The segments that contain a rhythm with length less than 3 seconds are discarded, because the period of the rhythm in such a segment may be too short for the detection. If we extract segments evenly on the recordings, then the segments of a single rhythm will be far more than that of multiple rhythms. However, this will lead to the results evaluated on these segments would not clearly reflect whether a model has the WSL ability to take advantage of the supervisory information of unspecified locations. Therefore, we take another method which extracts segments densely (in a stride of 5 seconds) in the rhythm transition region, but sparsely (in a stride of 250 seconds) in other regions. In this way, we get 1219 segments of pure AF rhythm, 1812 segments of other rhythms, and 1360 segments of both AF and other rhythms, as shown in Fig. 3(a).

As for the task of morphological arrhythmias detection, we utilize the MIT-BIH Arrhythmia Database (MITDB) [28, 29] for the evaluation. The rhythm type of each heartbeat in the database is annotated by cardiologists. As some types of heartbeat are very rare, here we only consider to detect 5 types of heartbeats: normal beat (N), left bundle branch block beat (LBBB), right bundle branch block beat (RBBB), atrial premature beat (APB) and premature ventricular contraction (PVC), according to a common practice in this domain [14, 30-32]. The recordings in this dataset are also split into segments of 20 seconds without overlap. Segments that don't contains heartbeat belonging to the above 5 classes are discarded, and the remaining segments are 4120 in total, as shown in Fig. 3(b). In our experiments, only the first channel of ECG in both datasets is used for model training and testing, although the processing of double-channel data is also supported by our design.

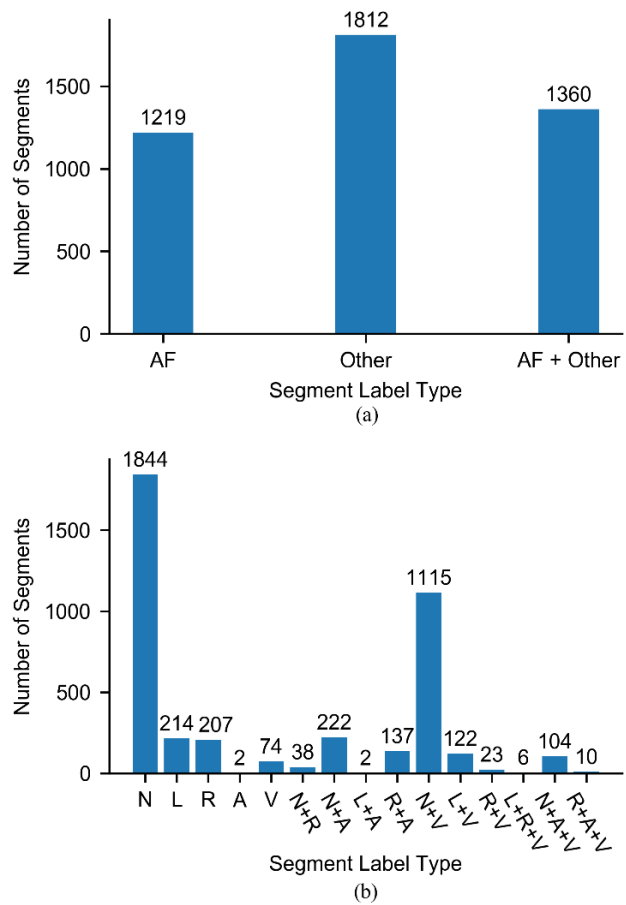

Fig. 3. Distributions of segment-level labels in the datasets used for cross validation in our experiments. (a) Distribution in the dataset extracted from AFDB. (b) Distribution in the dataset extracted from MITDB. The denotation of labels are as follows: "AF" denotes atrial fibrillation, "Other" denotes rhythms other than AF, "N" denotes normal rhythm, "L" denotes LBBB, "R" denotes RBBB, "A" denotes APB, and "V" denotes PVC. "A+B" indicates

## B. Metrics

The models for both detection tasks are evaluated in a 5-fold cross validation setting. The test results of all the folds are aggregated together for the metrics calculation. The metrics for detection of rhythmic and morphological arrhythmias are both calculated in beat level. The metrics used in our experiments includes sensitivity (*Se*) as in (7), specificity (*Sp*) as in (8), positive predictivity (*Ppr*) as in (9), and accuracy (*Acc*) as in (10).

$$Se = \frac{TP}{TP+FN}, \tag{7}$$

$$Sp = \frac{TN}{TN+FP}, \tag{8}$$

$$Ppr = \frac{TP}{TP+FP}, \tag{9}$$

$$Acc = \frac{TP+TN}{TP+FP+TN+FN}, \tag{10}$$

where, for a certain class, *TP*, short for true positive, is the number of correctly detected beats of the class; *FP*, short for false positive, is the number of beats that are falsely predicted as the class; *FN*, short for false negative, is the number of beats that belong to the class but are not detected by the model, and *TN*, short for true negative, is the number of correctly detected beats that don't belong to the class. For multi-class classification, we also calculate the overall value across all classes of each metric, which is the average weighted by the frequency of each class.

## C. Implementation of Models

In our experiments, the hyper-parameters of convolutional blocks are shared by all the tested models. For both types of blocks, each convolutional layer contains 32 filters, and the kernel size of each filter is 32. The weights of convolutional layers are initialized by the method proposed in [33]. The dropout rate of each dropout layer is 0.25 which is chosen experimentally. The depth and aggregation mechanism vary from model to model, which is discussed in particular in the following subsections. Besides, in each model, the part responsible for making local predictions consists of two convolutional layers: the number of filters in the first is 32, while that in the second is equal to the number of classes.

We implement the models based on the Keras framework [34] and train the models on a workstation with one CPU running at 3.5 GHz, an NVIDIA Quadro k6000 GPU, and 64 Gb of memory. The method used for model optimization is Adaptive Moment Estimation (Adam) [35], where $\beta_1$ is 0.9, $\beta_2$ is 0.999, and the learning rate is 0.001. The training of a model is terminated when no decrease of the loss value on the validation set is observed over 10 epochs.

## D. Assessment of Aggregation Mechanisms

The results of models with different aggregation mechanisms for AF detection are shown in Table I. All the models share the same base structure which consists of 7 convolutional blocks, but have different aggregation mechanisms. For each model, we report its performance on all the extracted 4391 segments and also that on only the 1360 rhythm transition segments which contain AF and other rhythms simultaneously. The ECG in transition regions may confuse the model and induce prediction errors, thus are particularly suitable for testing the performance of weak supervised learning. If a model can accurately identify the location of the rhythm transition, then it's all the more reason to believe that weak supervised learning does work. The results reported in the

TABLE I
Experimental Results on AF Detection with Different Aggregation Mechanisms

| Aggregation Mechanism | Performance on all segments | | | Performance on rhythm transition segments | | |
|---|---|---|---|---|---|---|
| | Se % | Ppr % | Acc % | Se % | Ppr % | Acc % |
| GAP | 86.89 | 95.15 | 91.47 | 71.98 | 90.12 | 80.17 |
| GMP | **95.71** | **96.51** | **96.23** | **91.35** | **93.46** | **91.69** |
| LSE (r = 3) | 88.83 | 96.07 | 92.81 | 76.20 | 91.64 | 83.03 |
| LSE (r = 5) | 85.31 | 95.29 | 90.82 | 69.43 | 91.26 | 79.45 |

table is the summary of the testing results in each fold, where we merge the confusion matrixes of all the folds into a total confusion matrix, and then compute the metrics based on the matrix. The same method is also used to compute the results in Table II, III, IV, VI, VII and IV.

From the results in Table I, we find that, among all the models, the model with GMP achieves the highest performance in each metric of the two segment sets. Specifically, the *Se* (95.71%) of this model is significantly higher than that of other models by over 8 percentage points on all segments and by over 15 percentage points on rhythm transition segments. For each model, its metrics evaluated on the rhythm transition segments are all lower than that on all segments, which confirms our previous inference that rhythm transition region is prone to misclassification. Nevertheless, the model with GMP exceeds 90% in terms of *Se*, *Ppr* and *Acc* on the rhythm transition segments. Therefore, GMP has a significant performance advantage compared with other evaluated aggregation mechanisms in the weakly supervised detection of AF.

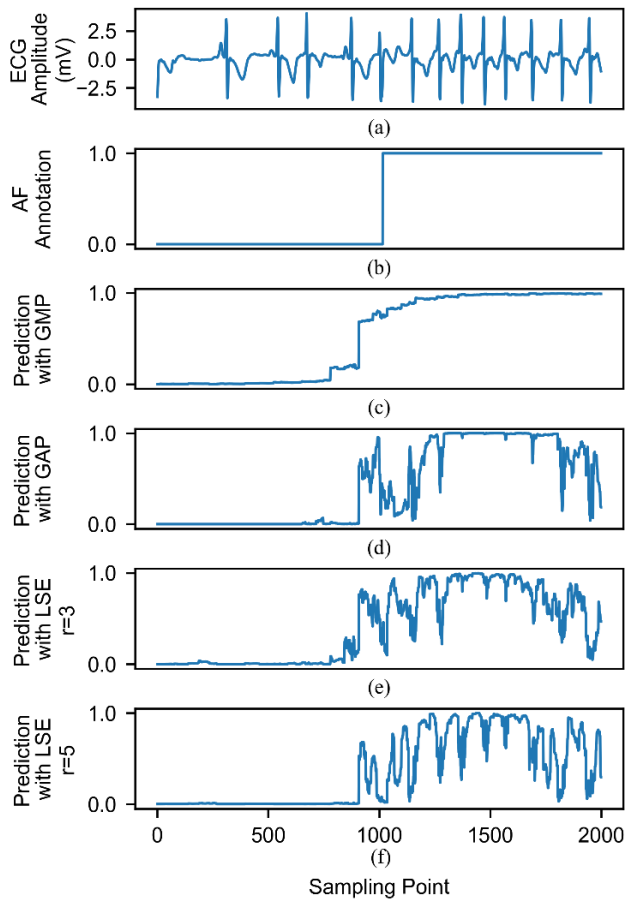

Fig. 4. Qualitative results of AF detection on the AFDB dataset. (a) ECG waveform. (b) The annotation of AF provided by the dataset. (c) The prediction of AF by the model with GMP. (d) The prediction of AF by the model with GAP. (e) The prediction of AF by the model with LSE where *r* is set to 3. (f) The prediction of AF by the model with LSE where *r* is set to 5.

In order to compare the influences of different aggregation methods on the local predictions more intuitively, we plot the local prediction maps on a rhythm transition (from other rhythms to AF) segment as shown in Fig. 4. The predictions of different models in the other rhythms period before the 800$^{th}$ sampling point are all stable and similar to each other. However, when time approaches the transition point of rhythm, the predictions of different models show significant differences: the AF probability predicted by the model with GMP is roughly monotonically increasing, while those predicted by other models fluctuate wildly. Then, in the AF period after the transition, the predictions of the model with GMP are still obviously more unstable than that of other models. It can be seen that the local prediction of the model with GMP is more in line with our expectation of a WSL model.

The experimental results for morphological arrhythmias detection are shown in Table II. The common base structure for these models consists of 5 convolutional blocks. The aggregation mechanisms evaluated for this task include GAP, GMP, LSER, and also their masked versions (i.e., GARP, GMRP and LSER). Among all the evaluated models, the model with GMRP achieves the highest performance in detecting all the 5 types of heartbeats with *Se* and *Ppr* both exceeding 95% for all the heartbeats except APB. Although its *Se* for APB is just 72.3%, it's still 17.6 points higher than that (54.7%) of the second-placed model, i.e., LSER (r=5). The models with masked aggregation mechanisms generally achieve higher *Se* in detecting APB and PVC than that with ordinary aggregation mechanisms. One possible explanation is that the fragments manifesting APB or PVC usually occupy only a small fraction of an ECG signal, thus they tend to be overwhelmed by the predictions of other parts when using ordinary aggregation methods.

TABLE II
Experimental Results on Morphological Arrhythmias Detection with Different Aggregation Mechanisms

| Aggregation Mechanism | N | | LBBB | | RBBB | | APB | | PVC | |
|---|---|---|---|---|---|---|---|---|---|---|
| | *Se*% | *Ppr*% | *Se*% | *Ppr*% | *Se*% | *Ppr*% | *Se*% | *Ppr*% | *Se*% | *Ppr*% |
| GAP | 97.0 | 96.6 | 97.9 | 97.2 | 98.1 | 82.9 | 25.4 | 37.7 | 67.3 | 77.4 |
| GMP | 99.1 | 87.6 | 10.9 | 99.2 | 34.5 | 89.3 | 22.7 | 91.2 | 59.3 | 38.1 |
| LSE (r = 3) | 96.0 | 97.8 | 97.8 | 96.4 | 97.6 | 84.2 | 31.5 | 47.1 | 81.0 | 71.6 |
| LSE (r = 5) | 95.9 | 97.2 | 97.1 | 97.9 | 97.2 | 83.0 | 28.3 | 34.9 | 76.5 | 73.5 |
| GARP | 87.2 | 98.4 | 93.0 | 99.2 | 91.5 | 90.7 | 50.6 | 33.8 | 92.3 | 44.2 |
| GMRP | **99.6** | **99.1** | **99.7** | **99.8** | **99.2** | **95.5** | **72.3** | **91.7** | **95.3** | **97.0** |
| LSER (r = 3) | 91.2 | 98.7 | 93.1 | 98.4 | 89.9 | 87.6 | 39.8 | 33.1 | 94.8 | 53.8 |
| LSER (r = 5) | 94.5 | 99.0 | 96.0 | 99.1 | 93.8 | 91.2 | 54.7 | 50.1 | 96.3 | 66.2 |

The local predictions by different models on a multi-label segment are shown in Fig. 5 (using ordinary aggregation mechanisms) and Fig. 6 (using masked aggregation mechanisms). We find that the predictions of the model with GMP, as in Fig. 5(b), at the R peaks are not very differentiated between the categories, but its predictions in the area immediately after the R peaks are more differentiated. It reflects that the predictions of the model for the beats are not well aligned with the R peaks, which may be an important reason for the poor performance of this model. By contrast, the predictions of models with GAP or LSE are well differentiated most of the time, and have good identification ability for RBBB. However, their predicted positions of APB are distributed over almost every heartbeat cycle, but the true location is not identified. This suggests that these models failed to recognize the real characteristics of APB, which are also reflected in their ultra-low scores in detecting APB as reported in Table II.

For the models with masked aggregation mechanisms, their predictions their predictions only make sense around the R peaks, because predictions elsewhere are masked out before aggregation during the training process. In consequence, we only need to look at their predictions around the R peaks, which are zoomed in horizontally for the convenience of observation as shown in the dashed boxes in Fig. 6. The model with GMRP gives very differentiated predictions at each R peak of the segment, which is in sharp contrast to the predictions of the model with GMP. The predictions of other models in Fig. 6 exhibit similar patterns to that of the model with GMRP, but with ambiguities or errors at some R peaks. Therefore, compared with other aggregation mechanisms, GMRP is more effective for the weakly supervised detection of morphological arrhythmias.

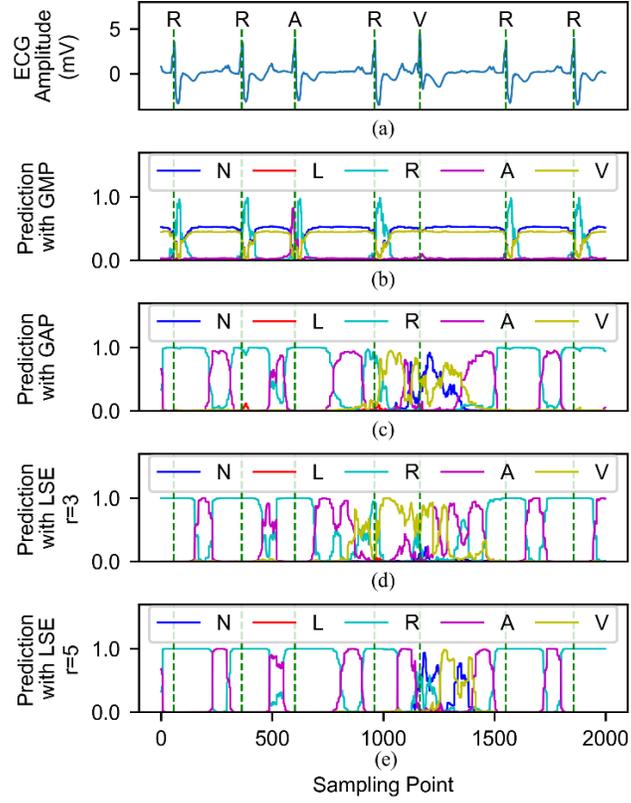

Fig. 5. Qualitative results of morphological arrhythmias detection on the MITDB dataset using ordinary aggregation mechanisms that are not discriminatory for positions. (a) ECG waveform, where R peak positions are marked by vertical dashed green lines, and the ground truth of beat type is labeled above each R peak. (b) The prediction of arrhythmias by the model with GMP. (c) The prediction of arrhythmias by the model with GAP. (d) The prediction of arrhythmias by the model with LSE where *r* is set to 3. (e) The prediction of arrhythmias by the model with LSE where *r* is set to 5. The denotation of labels are as follows: "N" denotes normal rhythm, "L" denotes LBBB, "R" denotes RBBB, "A" denotes APB, and "V" denotes PVC.

### E. Relevance of Model Depth

In this subsection, we evaluate the influence of model depth to the performance of proposed WSL models. As the main body of our network consists of convolutional blocks, we measure the model depth by the number of convolutional blocks in the network. In theory, model depth can affect the performance of a model in at least two ways. On one hand, model depth is positively correlated with the model complexity and also their ability to abstract data. On the other hand, the size of the receptive field is also affected by the model depth: usually the deeper the model, the larger the receptive field.

In our experiments, 5 models of depth from 5 to 9 blocks are trained to detect AF, with the results shown in Table III. The performance of these models is similar to that of each other, except for the model with 5 blocks whose *Se* and *Acc* are obviously lower than that of other models. It indicates that, for the weakly supervised detection of AF, a too shallow model can lead to significant performance degradation, while as the model deepens gradually and reaches a certain depth, its performance tends to be stable. One possible reason for this is that the receptive field of a too shallow model is not sufficient to extract the solid features to detect AF, in view of that a rhythmic abnormality usually tasks a little longer (e.g., several seconds) to fully manifest.

The results of models of different depths for the detection of morphological arrhythmias are shown in Table IV. The model with 5 blocks achieves the best performance in detecting each of the 5 types of beats. Both the deeper or shallower models show some performance degradation, especially for the detection of APB. This tells us that, for the weakly supervised detection of abnormal heartbeats, the model depth is critical to the performance of a model. It makes logical sense, since the features of a heartbeat is usually manifested in a relatively fixed time scale, and too large or too small a receptive field can affect the extraction of its features.

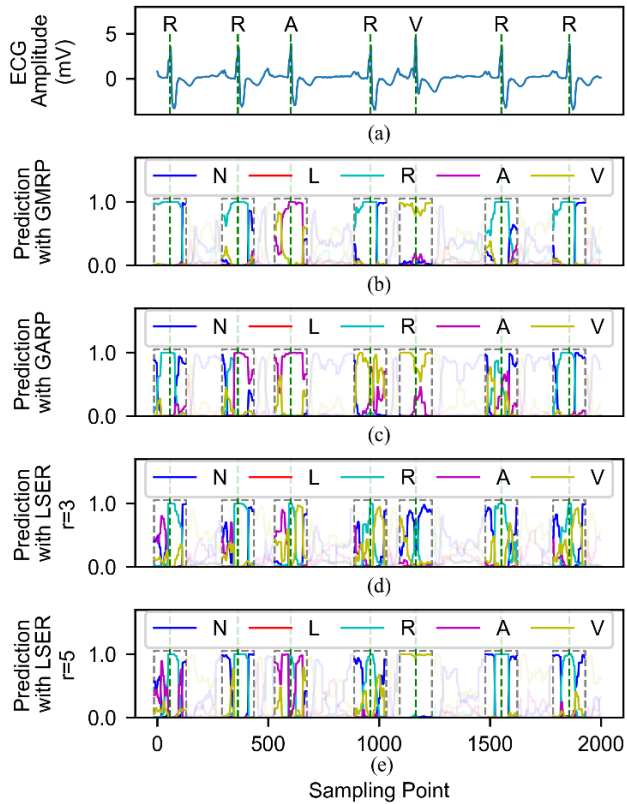

Fig. 6. Qualitative results of morphological arrhythmias detection on the MITDB dataset using masked aggregation mechanisms that only aggregate the predictions on R peaks. (a) ECG waveform and labels of true beat types. (b) The prediction of arrhythmias by the model with GMRP. (c) The prediction of arrhythmias by the model with GARP. (d) The prediction of arrhythmias by the model with LSER where $r$ is set to 3. (e) The prediction of arrhythmias by the model with LSER where $r$ is set to 5. For the convenience of observation, we zoom in 4 times horizontally around the R peaks as shown in the dashed gray boxes. The parts outside the dashed boxes are reduced in opacity to avoid distracting the observers.

TABLE III
Experimental Results on AF Detection with Different Network Depth

| Depth (blks) | Performance on all segments | | | Performance on rhythm transition segments | | |
|---|---|---|---|---|---|---|
| | *Se* % | *Ppr* % | *Acc* % | *Se* % | *Ppr* % | *Acc* % |
| 5 | 82.74 | 96.53 | 90.17 | 78.37 | 93.72 | 85.16 |
| 6 | 95.66 | 96.09 | 95.98 | **91.40** | **93.74** | **91.88** |
| 7 | 95.71 | 96.51 | **96.23** | 91.35 | 93.46 | 91.69 |
| 8 | 95.64 | **96.55** | 96.22 | 90.78 | 93.32 | 91.32 |
| 9 | **95.76** | 96.32 | 96.16 | 91.25 | 92.49 | 91.08 |

"blks", short for blocks, is the unit of measurement for the model depth.

TABLE IV
Experimental Results on Morphological Arrhythmias Detection with Different Network Depth

| Depth (blks) | N | | LBBB | | RBBB | | APB | | PVC | |
|---|---|---|---|---|---|---|---|---|---|---|
| | Se % | *Ppr*% | Se % | *Ppr* % | Se % | *Ppr* % | Se % | *Ppr* % | Se % | *Ppr* % |
| 4 | 99.5 | 98.9 | 99.1 | 99.4 | 98.6 | 92.9 | 58.7 | 87.3 | 94.1 | 94.7 |
| 5 | **99.6** | **99.1** | **99.7** | **99.8** | **99.2** | **95.5** | **72.3** | **91.7** | **95.3** | **97.0** |
| 6 | 99.3 | 98.9 | 99.3 | 99.6 | 99.2 | 88.0 | 39.9 | 86.5 | 94.5 | 92.9 |
| 7 | 98.6 | 97.1 | 79.2 | 99.4 | 95.3 | 84.0 | 42.1 | 88.3 | 91.9 | 82.3 |

14## F. Influence of Data Distribution

Data distribution is another important factor to be considered when evaluating a WSL model. In the previous subsection, we have seen the impacts of different distributions of testing data (all segments vs. rhythm transition segments) on the model performance, now we examine the effects of different distributions of the training data. We consider the data distributions in the segment level, i.e., the distribution of segment labels which are the sources of the supervision information available to the models. The original distributions of segments extracted from AFDB and MITDB are shown in Fig. 3. We classify the segments into 2 types: single-label segment and multiple-label segment. A single-label segment has only one type of label in its annotation, e.g., AF or LBBB, while a multiple-label segment is annotated by a combination of several labels, e.g., AF+Other or N+APB. Each of our original segment sets contains these two types of segments simultaneously. However, for WSL, the intensity of supervision provided by these two kinds of segments is different. Theoretically speaking, the more labels a segment has, the more possible permutations of the labels in its time period, and thus the weaker supervision it provides. If a dataset contains only single-label segments, the problem actually departs into full supervised learning. Therefore, by changing the data distribution, we can better understand the real effect of WSL.

Here, we propose 4 distribution modifications as shown in Table V. The basic idea behind these modifications is to change the ratio between different types of segments by reducing the number of segments of a certain type. To facilitate the comparison of effects across distributions, in each fold of the cross validation, we adjust only the training set and the validation set, but leave the testing set unchanged. Among these modifications, Single50 and OnlyMultiple reduce the number of single-label segments to 50% and 0% of their original, respectively. Similarly, Multiple50 and OnlySingle remove 50% and 100% of multiple-label segments, respectively.

TABLE V
Modifications of Segments Distribution in Each Fold

| Modified Distribution | Training Set | | Validation Set | | Testing Set | |
|---|---|---|---|---|---|---|
| | SL% | ML% | SL% | ML% | SL% | ML% |
| Single50 | 50 | 100 | 50 | 100 | 100 | 100 |
| Multiple50 | 100 | 50 | 100 | 50 | 100 | 100 |
| OnlySingle | 100 | 0 | 100 | 0 | 100 | 100 |
| OnlyMultiple | 0 | 100 | 0 | 100 | 100 | 100 |

"SL", short for single label, denotes segments with only one label. "ML", short for multiple labels, denotes segments with multiple labels. Each figure in this table represents the percentage of the corresponding segments retained in the corresponding dataset after the modification.

Three modified data distributions are used to evaluate the performance of AF detection with the results shown in Table VI. OnlyMultiple is not used for this evaluation, because its every segment has the same label combination (i.e., AF+Other) and thus cannot provide any supervision information for the model. The model to evaluate consists of 7 convolutional blocks and uses GMP as the aggregation mechanism, and its results on the original dataset have also been reported in Table II and III. From Table VI, we see that the performance of the model trained on Multiple50 is obviously lower than that on the original dataset, while that on OnlySingle is even worse. It indicates that the model does use the weak supervisory information from the multiple-label segments

TABLE VI
Experimental Results on AF Detection with Different Data Distributions of the Training Set

| Data Distribution | Performance on all segments | | | Performance on rhythm transition segments | | |
|---|---|---|---|---|---|---|
| | *Se* % | *Ppr* % | *Acc* % | *Se* % | *Ppr* % | *Acc* % |
| Original | **95.71** | **96.51** | **96.23** | **91.35** | **93.46** | **91.69** |
| Single50 | 94.56 | 95.85 | 95.37 | 89.69 | 92.74 | 90.44 |
| Multiple50 | 94.42 | 96.30 | 95.53 | 88.43 | 92.97 | 89.92 |
| OnlySingle | 89.11 | 96.07 | 92.94 | 75.07 | 91.08 | 82.18 |

TABLE VII

Experimental Results on Morphological Arrhythmias Detection with Different Data Distributions of the Training Set

| Data Distribution | N | | LBBB | | RBBB | | APB | | PVC | |
|---|---|---|---|---|---|---|---|---|---|---|
| | Se% | PPv% | Se% | Ppr% | Se% | Ppr% | Se% | Ppr% | Se% | Ppr% |
| Original | 99.6 | **99.0** | 99.4 | 99.5 | 99.1 | 95.1 | 69.1 | 91.7 | 94.4 | 96.2 |
| Single50 | 99.0 | 98.7 | 94.5 | 99.2 | 98.4 | 94.2 | 55.8 | 84.7 | **94.5** | 85.8 |
| Multiple50 | 99.4 | 98.4 | 99.2 | 99.2 | 97.3 | 92.4 | 50.8 | 82.4 | 91.9 | 94.5 |
| OnlySingle | **99.8** | 96.4 | 97.8 | 74.3 | 98.0 | 80.5 | 0 | 0 | 39.7 | **96.9** |
| OnlyMultiple | 78.9 | 97.5 | 34.7 | 94.2 | 81.0 | 90.3 | 61.5 | 19.1 | 87.2 | 28.7 |

to optimize its parameters. The model trained on Single50 also show some performance degradation which reflects the important role of single label samples in the model training.

For the detection of morphological arrhythmias, all the 4 modified distributions are used to evaluate the model's performance. The model used in these experiments consists of 5 convolutional blocks, and use GMRP as the aggregation mechanism. Table VII shows the results of these experiments. Compared with the model trained on the original dataset, the model trained on Multiple50 shows slight performance declines in detecting RBBB and PVC, and shows a greater decline in detecting APB. When all the multiple-label segments are removed from the training set, more significant declines are seen in the *Ppr* for RBBB and LBBB, and the *Se* for PVC. The model trained on OnlySingle is not suitable for the APB detection, since there are only 2 single-label APB segments. When the single-label segments in the training set are subtracted by half, the Se for RBBB, Ppr for PVC, and both the metrics for APB are obviously lower than that of the original model. Furthermore, when the model is trained with only multiple-label segments, its detection performance for all the categories decreases significantly. Therefore, at least for this dataset, both the single-label segments and multi-label segments play important roles in improving the performance of the model for weakly supervised arrhythmias detection.

*G. Comparison with Other Works*

To further validate the performance of the method, we compare the results with that of other studies, although they are either manually designed algorithms or fully supervised learning methods. Based on the previous analysis, two models are selected for the comparison with other studies: the model with 7 convolutional blocks and GMP is trained to detect AF, while the model with 5 convolutional blocks and GMRP is trained for the detection of morphological arrhythmias. Because most studies for AF detection based on AFDB reported their results on the whole dataset or on a random subset of the dataset, we also report the results of testing our model across the whole AFDB dataset, as shown in Table VIII. The model is trained in one fold of the cross validation as stated above, with the training data accounting for about 6% of the whole dataset. As for the model for morphological arrhythmias detection, its confusion matrix is an aggregation of the smaller confusion matrixes of each fold in the cross validation, and then is used to compute the scores on each metric, because the whole MITDB dataset has been used in the cross validation. The results of this model are shown in Table IX, where the overall values (i.e., the averages weighted by the frequency of each class) of the metrics are present on the bottom line.

TABLE VIII

Performance of the Suggested Model for AF Detection on the Whole AFDB

| | | Pred | | Se % | Sp % | Ppr % | Acc % |
|---|---|---|---|---|---|---|---|
| | | AF | Other | | | | |
| Ref | AF | 252423 | 2988 | 98.83 | 99.37 | 99.38 | 99.09 |
| | Other | 1566 | 245345 | 99.37 | 98.83 | 98.80 | 99.09 |

The results of different studies for AF detection are shown in Table X. The methodologies adopted by these studies range from manually designed algorithms [36-38], traditional machine learning methods [39, 40], to DNN-based methods [5, 15, 16, 41]. In particular, all the machine learning based methods are fully supervised. We note that different studies usually have different prediction units, i.e., the resolution of their predictions in time. For example, the method proposed by Xia et al. [15] makes

TABLE IX

Performance of the Suggested Model for Morphological Arrhythmias Detection on the Whole MITDB

|     |      | Pred  |      |      |      |      | Se % | Sp % | Ppr % | Acc % |
|-----|------|-------|------|------|------|------|------|------|-------|-------|
|     |      | N     | LBBB | RBBB | APB  | PVC  |      |      |       |       |
| Ref | N    | 74171 | 10   | 14   | 105  | 146  | 99.63 | 97.07 | 99.03 | 98.99 |
|     | LBBB | 30    | 7978 | 0    | 0    | 19   | 99.39 | 99.95 | 99.46 | 99.91 |
|     | RBBB | 24    | 0    | 7154 | 30   | 9    | 99.13 | 99.60 | 95.13 | 99.57 |
|     | APB  | 329   | 4    | 348  | 1726 | 92   | 69.07 | 99.84 | 91.71 | 99.06 |
|     | PVC  | 345   | 29   | 4    | 21   | 6686 | 94.37 | 99.71 | 96.17 | 99.33 |
|     |      |       | Overall |   |      |      | 98.43 | 97.74 | 98.39 | 99.13 |

The confusion matrix is the aggregation of the sub-matrixes that are computed based on the test data of the corresponding folds in the cross validation.

predictions in the unit of 5-second segments, while the method proposed by Lee et al. [38] needs a 128-beat segment to determine whether or not there is an AF episode. As a result, when comparing different studies, we need to pay attention to their prediction units, which have been annotated in Table X. It is obvious that a smaller prediction unit is more desirable because it can bring a higher prediction resolution. Actually, beat-level resolution is sufficient in the clinical condition, thus we use 1 beat as the prediction unit of our model to evaluate its performance. As the results show, our model achieves the best *Sp* among these studies, and its *Se*, *Ppr* and *Acc* are also very close to the corresponding best ones. These results are noteworthy, especially since we use a WSL approach and a smaller prediction unit.

Table XI shows the results of different methods that are tested on the WFDB dataset for the detection of beat-level arrhythmias. Except the study by Martis et al. [31] which adopts the traditional machine learning method, other studies all adopt DNN-based methods. A wide spectrum of network structures have been used for this task, ranging from CNN [32], faster R-CNN [42], U-net [14], LSTM [43, 44] to auto-encoder [7, 44]. Many other studies that have sought to achieve similar goals are not reported in this table, because they use different datasets or categories to classify. Although all the reported studies use the same dataset and has

TABLE X

Performance of Published Other Methods for AF Detection

| Authors | Year | Method | Unit | Se % | Sp % | Ppr % | Acc % |
|---------|------|--------|------|------|------|-------|-------|
| Huang et al. [36] | 2011 | delta RR interval distribution difference analysis | 1 beat | 96.1 | 98.1 | N/A | N/A |
| Jiang et al. [37] | 2012 | RR intervals and P wave analysis | 1 beat | 98.2 | 97.5 | N/A | N/A |
| Lee et al. [38] | 2013 | Time-Varying Coherence Function + Shannon entropy | 128 beats | 98.2 | 97.8 | N/A | N/A |
| Asgari et al. [39] | 2015 | Stationary wavelet transform + support vector mechine | 30 seconds | 97.0 | 97.1 | N/A | 97.1 |
| Xia et al. [15] | 2017 | Stationary wavelet transform + CNN | 5 seconds | 98.79 | 97.87 | N/A | 98.63 |
| He et al. [6] | 2018 | Continuous wavelet transform + CNN | 5 beats | 99.41 | 98.91 | 99.39 | 99.23 |
| Kalidas et al. [40] | 2019 | discrete-state Markov models + Random Forests | 60 seconds | 97.4 | 98.6 | N/A | N/A |
| Lai et al. [41] | 2019 | RR intervals + F-wave frequency spectrum + CNN | 10 seconds | 97.8 | 97.2 | N/A | 97.5 |
| Mousavi et al. [5] | 2019 | CNN + Attention | 5 seconds | 99.53 | 99.26 | N/A | 99.40 |
| **Current** | **2020** | **WSL (ResNet + GMP)** | **1 beat** | **98.83** | **99.36** | **99.38** | **99.09** |

"N/A" indicates the data is not available.

TABLE XI

Performance of Published Other Methods for Morphological Arrhythmias Detection

| Authors | Year | Method | Beat distribution | | | | | Se % | Sp % | Ppr % | Acc % |
|---------|------|--------|---|---|---|---|---|------|------|-------|-------|
|         |      |        | N | LBBB | RBBB | APB | PVC |   |   |   |   |
| Martis et al. [31] | 2013 | LS-SVM with RBF kernel | 10000 | 8069 | 7250 | 2544 | 7126 | 99.27 | 98.31 | 99.33 | 93.48 |
| Li et al. [32] | 2017 | CNN | 2640 | 2973 | 3000 | 3000 | 1200 | 97.42* | 99.33* | 97.43* | 98.88* |
| Oh et al. [43] | 2018 | CNN + LSTM | 8245 | 344 | 660 | 1004 | 6246 | 97.52* | 98.41* | 97.49* | 98.55* |
| Yildirim et al. [44] | 2019 | Convolutional auto-encoder + LSTM | 11199 | 1187 | 1101 | 406 | 1111 | 99.11* | 98.66* | 99.10* | 99.38* |
| Oh et al. [14] | 2019 | Modified U-net | 71337 | 7890 | 7123 | 2123 | 6194 | 96.13* | 95.84* | 96.05* | 97.70* |
| Ji et al. [42] | 2019 | Faster R-CNN | 5250 | 5037 | 4629 | 0 | 4564 | 98.06 | 99.45 | 98.04* | 99.21 |
| Hou et al. [7] | 2020 | LSTM-based auto-encoder + SVM | 74749 | 8071 | 7255 | 2546 | 7123 | 99.35 | 99.84 | N/A | 99.74 |
| **Current** | **2020** | **WSL (ResNet + GMRP)** | **74446** | **8027** | **7217** | **2449** | **7085** | **98.43** | **97.74** | **98.39** | **99.13** |

* indicates the figure is computed by us from the confusion matrix provided in the source paper.

almost the same classification goals, the beat-level distributions of their testing set are different from each other. Some studies tested their methods on a subset of the original dataset, and some studies even changed the ratio between sample sizes for each category. These changes will undoubtedly have impacts on the results, so we present the distribution of testing set of each study in table XI to get the reader's attention. Besides, due to some studies didn't provide their results on some metrics, we calculate these results, marked with "*" in the table, by ourselves based on the confusion matrixes provided in their paper. The results show that the performance of our model is comparable to that of the fully supervised methods. Among these studies, Oh et al. [14] and Hou et al. [7] use almost the same testing set as ours, thus the comparison between these three studies makes more sense. Our model ranks the second on each metric among the three studies, although we only use a weakly supervised methodology.

## V. Discussion

The experimental results demonstrate the efficiency of the proposed weakly supervised methods for the detection of AF and morphological arrhythmias. The framework essentially uses the global labels as a constraint on local predictions and optimizes the local predictions indirectly by optimizing the performance of global predictions. When the sample size of the training set is large enough, the parameters of the model will converge to a small range, and stable predictions will be given. The aggregation mechanism in the framework is critical especially when the sample size is not enough to make the model parameters fully convergent. So, we see that the masked aggregation mechanisms significantly improved the detection performance for beat-level arrhythmias, since they reduced the number of possible solutions for local predictions. Considering that there are a large number of simply labeled ECG data in clinical practice, this technique is expected to play an important role in the exploration and utilization of these data.

Another advantage of our method is that it can process input of any length at a stroke, and thus are more efficient than methods that make predictions in sliding windows. Generally, in order to make the predictions more detailed in time scale, many methods detect arrhythmias in small windows that slide along time. However, the sliding windows, usually overlapping with its neighbors, can not only lead to greater computational overhead, but also break the context required for the predictions. By contrast, our method can make detailed predictions for input of any length, hence avoids the drawbacks of sliding window.

This method can also play an important role to interpret and diagnose the model even if detailed local predictions are not the goals. Recall the example shown in Fig. 4, the model with GAP makes some wrong predictions after the rhythm transition point, which reveals its weakness to recognize some features of AF. By using this method, researchers can locate the defects of a model more quickly and find targeted solutions.

The proposed method also has its limitations especially on the detection of morphological arrhythmias. The masked aggregation mechanisms depend on the positions of QRS complexes. Although many methods have achieved very high accuracy for the QRS detection on the MITDB, their performance may degrade when the signal quality is very poor. Therefore, the dependence on the third-party QRS detection algorithms will bring uncertainty to the performance of our model. The imbalance between categories is another problem that needs to be further addressed, even for fully supervised learning. The low performance for the APB detection is partly due to its small sample size compared with other categories. In future studies, we will explore new methods to enhance the model's performance in dealing with low quality signals and imbalanced categories.

## VI. Conclusion

In this work, we proposed a WSL framework for the detection of both rhythmic and morphological arrhythmias. Our model only requires record-level annotations as the supervision during training but can efficiently generate detailed local predictions for the testing data with the performance comparable to that of the fully supervised methods. We revealed the influence of aggregation mechanisms, model depths and data distributions on the performance of our method through experiments. In particular, the masked aggregation mechanism plays an important role to improve the performance for the detection of morphological arrhythmias. Our model has great potential in exploiting the existing ECG recording data, reducing the cost of manual labeling and diagnosing the DNN models.

Acknowledgment